\pdfoutput=1

\documentclass[11pt]{article}

\usepackage[final]{acl}

\usepackage{times}
\usepackage{latexsym}

\usepackage[T1]{fontenc}

\usepackage[utf8]{inputenc}

\usepackage{microtype}

\usepackage{inconsolata}

\usepackage{graphicx}

\usepackage{booktabs}
\usepackage{subcaption}
\usepackage{blindtext}
\usepackage{hyperref}
\usepackage{csquotes}
\usepackage{amsmath}
\usepackage{array}

\title{\textsc{SceneGram}: Conceptualizing and Describing Tangrams in Scene Context}

\author{Simeon Junker \and Sina Zarrieß \\
  Computational Linguistics, Department of Linguistics\\
  Bielefeld University, Germany\\
  \texttt{\{simeon.junker,sina.zarriess\}@uni-bielefeld.de}\\
  }

\begin{document}
\maketitle
\begin{abstract}
Research on reference and naming suggests that humans can come up with very different ways of conceptualizing and referring to the same object, e.g. the same abstract tangram shape can be a \enquote{crab}, \enquote{sink}, or \enquote{space ship}.
Another common assumption in cognitive science is that scene context fundamentally shapes our visual perception of objects and conceptual expectations.
This paper contributes \textsc{SceneGram}, a dataset of human references to tangram shapes placed in different scene contexts, allowing for systematic analyses of the effect of scene context on conceptualization.
Based on this data, we analyze references to tangram shapes generated by multimodal LLMs, showing that these models do not account for the richness and variability of conceptualizations found in human references.\footnote{Data and code for this project are available at: \href{https://github.com/clause-bielefeld/scenegram}{github.com/clause-bielefeld/scenegram}}
\end{abstract}

\section{Introduction}

\begin{figure*}
    \footnotesize
    \centering
    \begin{tabular}{p{0.3\textwidth}p{0.3\textwidth}p{0.3\textwidth}}
        \centering\arraybackslash \includegraphics[width=.22\textwidth]{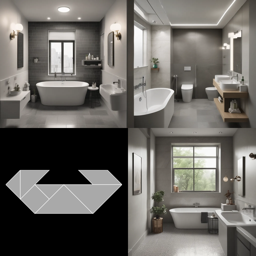} &
        \centering\arraybackslash \includegraphics[width=.22\textwidth]{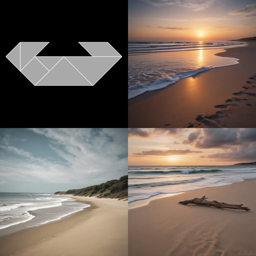} &
        \centering\arraybackslash \includegraphics[width=.22\textwidth]{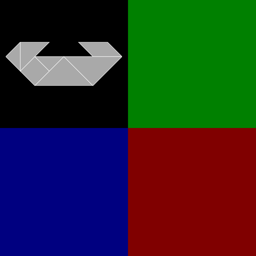} \\
        \textbf{human}: sink (5); bowl (2); crab (2); bathtub shape (1)
        & \textbf{human}: crab (7); bathtub (1); bowl (1); bull (1)
        & \textbf{human}: crab (4); bowl (2); dog (1); seal (1); letter c (1); space ship (1)\\
        \textbf{LLaVA 7b}: bathtub (6); rectangle (2); bathroom (2) &
        \textbf{LLaVA 7b}: sun (3); bird (2); diamond (2); boat (1); wave (1); house (1) &
        \textbf{LLaVA 7b}: house (3); square (2); diamond (2); triangle (1); parallelogram (1); box (1) \\
        \textbf{LLaVA 72b}: house (8); boat (1); bathtub (1)&
        \textbf{LLaVA 72b}: sailboat (4); house (3); boat (3)&
        \textbf{LLaVA 72b}: house (7); boat (3)\\
    \end{tabular}
    
    \caption{A single tangram in \textit{bathroom}, \textit{beach} and baseline contexts with counts for labels in annotated or predicted descriptions. 
    Human annotations commonly include labels which are coherent with scenes (\enquote{sink} and \enquote{crab}). LLaVA 7b and 72b show similar patterns, but also issues in differentiating between tangrams and scenes.}
    \label{fig:item_example}
\end{figure*}

Reference to visual objects is an elementary component of language and situated interaction. 
Almost always, we can refer to something in many different ways, and the linguistic choices we make reflect the ways we categorize or conceptualize it. More often than not, we can have multiple \textit{conceptual perspectives} \citep{Clark1997} on the same things:
For example, the same person could be referred to as a \enquote{human}, \enquote{woman}, \enquote{engineer}, \enquote{mom} or using their proper name, depending on what is deemed relevant, appropriate or useful in a given situation (e.g. \citealt{Brown1958, Graf2016}). 
Even fewer limitations exist for abstract shapes such as \textit{tangrams}:
In Figure \ref{fig:item_example}, the same shape can be seen as representing e.g., a \enquote{crab}, \enquote{sink}, or \enquote{space ship}, showcasing the flexibility and richness of human conceptualization and interpretation at the intersection of visual and semantic processing.
Despite recent progress in multimodal language modeling \citep{zhang-etal-2024-mm}, current systems show mixed results in reproducing human variation in object naming \citep{testoni-etal-2024-naming} and figurative descriptions for abstract stimuli remain a major challenge in vision and language (V\&L) research \citep{Ji2022, gul-artzi-2024-cogen}.

How can we navigate this complex web of many-to-many relationships between visual stimuli and possible conceptualizations and descriptions?
A common assumption in linguistics, cognitive science, and psychology is that our cognition is highly \enquote{tuned} to the everyday contexts and situations we interact in. A well-researched instance of this is our visual perception of objects which is known to be fundamentally shaped by high-level, conceptual expectations on the level of scenes (\citealt{Biederman1982, Greene2013, Vo2021}, among many others). 
For example, at the beach, we would rather expect to see certain animals and plants than household items, whereas in a bathroom, it would be the other way around.
It is well-researched that these processes facilitate, e.g., visual object recognition (e.g. \citealt{Bar2004}), and it seems plausible that scene context also affects descriptions of objects which can be conceptualized in different ways. 
However, existing V\&L datasets with real-world images make it difficult to study the same objects in different scenes:
As objects often occur in typical contexts, the same regularities that are exploited for visual processing obstruct investigation with natural data.

In this paper, we contribute (i) a dataset of human references to tangram shapes placed in different scene contexts, allowing for systematic analyses of the effect of scene context on the conceptualization of the same shape, and (ii) analysis of references to tangram shapes generated by multimodal LLMs, showing that these models do not account for the richness and variability of conceptualizations found in human references.
Our findings show that scene context affects tangram descriptions in a way, that speakers tend to verbalize conceptualizations that are consistent with the provided scene context. Experiments with MLLMs show similarities but also shed light on limitations and biases relevant to various topics in vision and language research.

\section{Background}

\paragraph{Human scene understanding}

Research on human vision and perception has shown that when viewing a scene, humans perceive it as a coherent whole instead of mere collections of objects \citep{Vo2021}. Capturing the \textit{gist} of a scene is a rapid process \citep{Oliva2006}, and while incongruent context can also be misleading \citep{Zhang2020a, Gupta2022}, scene-level information has been demonstrated to facilitate e.g., visual object recognition in both human cognition (\citealt{Palmer1975,Oliva2007,Parikh2012,Lauer2018}, among others) as well as computer vision (\citealt{Divvala2009,Galleguillos2010}, see \citealt{Wang2023} for a survey) or reference generation systems \citep{junker-zarriess-2024-resilience-scene}. 
For this, humans and machines can exploit learned knowledge about regularities of the visual word for visual processing \citep{Biederman1972,Bar2004,Greene2013,Pereira2014,Sadeghi2015}, e.g. \textit{semantic rules} that certain objects tend to occur in some contexts rather than others \citep{Biederman1982,Vo2021,Turini2022}.

\paragraph{Context, Conceptual Perspective and Referential Choice}

Verbal reference to visual objects requires making linguistic choices, as the same things can be called and described in many different ways \citep{Brown1958, Graf2016, Davies2019}. Research in Referring Expression Generation has modeled these choices as a function of \textit{context} \citep{Schuez2023}, i.e., objects co-occurring with the target are factored in to determine which properties have to be realized to make a description unambiguous in a given scenario (see \citealt{Krahmer2012} for a survey).
More generally, however, speakers can take on different \textit{conceptual perspectives} on referents, highlighting different (often not mutually exclusive) facets and aspects of referents, guided by principles beyond pragmatic informativeness \citep{Clark1997a, Gatt2006, Gatt2007a, Gatt2007b, Deemter2016}. Importantly, different conceptualizations can be reflected in object labels or names \citep{Clark1997, Gualdoni2023}, which have been shown to be highly varied and flexible \citep{Ordonez2016, Zarriess2017, Silberer2020, Silberer2020a, Gualdoni2022a, gualdoni-etal-2022-horse, Gualdoni2023}. 
Recent work in vision and language research has started to model this variation \citep{Ilinykh2023,testoni-etal-2024-naming}, but general questions remain about how visual context affects conceptualization and naming in humans and generation systems.

\paragraph{Tangrams in linguistic research}

Tangrams are abstract figures, which are constructed from a small set of geometric primitives and can be more or less \textit{nameable}, i.e., easy or hard to describe \citep{Zettersten2020}. Unlike natural objects, tangrams lack established naming conventions, triggering diverse figurative descriptions and making them suitable as stimuli to investigate linguistic reference in humans \citep{Clark1986, Schober1989, WilkesGibbs1992, Brennan1996, Murfitt2001, Hawkins2020a, Bangerter2020, Fasquel2022, Sudo2022} and vision-language systems \citep{Skantze2022, Ji2022, gul-artzi-2024-cogen}. \citet{Shore2018, Ji2022} released crowdsourced datasets using tangram figures as stimuli, we use tangrams from the latter for our work.

\paragraph{Research Gap}

While e.g. \textsc{ManyNames} \citep{Silberer2020, Silberer2020a} quantifies naming variation for objects in photographs, object types are often bound to certain contexts (reflecting real-life patterns), and most objects are highly nameable and easy to identify, limiting the range of different conceptualizations. In contrast to this, tangram datasets like \textsc{KiloGram} \citep{Ji2022} offer rich variation in conceptualizations, but do not account for contextual influences, as items are described in isolation.
In this work, we take a different approach and pair abstract tangram shapes with generated images representing a taxonomy of scene contexts. In this way, we collect diverse descriptions of visual items, which we subsequently analyze for context effects and compare with the predictions of multimodal LLMs.

\section{Data Collection}

We combine tangram figures with images depicting different types of scenes, and crowdsource annotations to investigate how context affects the conceptualization of inherently ambiguous shapes.

Formally, each item in our dataset $i \in I$ is defined as a tangram $t_i \in T$ with a scene $s_i \in S$ as visual context, i.e. a tuple $i = \langle t_i, s_i \rangle$. 
For each item, we collect a set of $D_i = D_{t_i, s_i}$ descriptions in English. From each annotated description, we extract an object label and the corresponding WordNet \citep{Miller1995} synset, to reduce onomasiological variation and facilitate taxonomy-based analyses. 

\subsection{Item Design and Generation}

As the tangrams for our dataset, we use half of the items from the \textit{dense} split in \textsc{KiloGram} (\citealt{Ji2022}, $|T|=37$), which come with rich annotations that can be used for comparison. 

Scene images ($|S|=11$) are generated for a set of scene categories using \textit{SDXL-Lightning} \citep{Lin2024} as a state of the art text-to-image model.\footnote{\href{https://huggingface.co/ByteDance/SDXL-Lightning}{huggingface.co/ByteDance/SDXL-Lightning}, accessed via the provided \href{https://huggingface.co/spaces/ByteDance/SDXL-Lightning}{Huggingface Space}} 
We generate three images for each of the 8 basic level scene categories in \citet{Lauer2018}, 
which include various indoor scenes (\textit{kitchen}, \textit{bathroom}, \textit{bedroom}, \textit{office}), for which we expect a particular influence due to their relation to common everyday objects, but also a broad selection of typical outdoor scenes (\textit{forest}, \textit{mountain}, \textit{beach}, \textit{street}). 
In addition to this, we include \textit{sky} and \textit{sea bottom} as additional outdoor scene categories, which are associated with certain objects that would be less expected in the remaining scenes (e.g., fish or birds).
We prompt the model to generate \enquote{a photograph of a \textsc{[scene]}}, where \textsc{[scene]} is replaced with the respective scene category label. We also add \textit{none} as the baseline scene condition $s_b$ with neutral context, i.e., uniform color patches. 

Our final items ($n = |T| \times |S| = 407$) combine tangrams and scene images by arranging them into a $2 \times 2$ grid of random order. Here, one tile is always occupied by a tangram shape and the remaining three tiles by images depicting a specific type of scene (cf. Figure \ref{fig:item_example}). This procedure is intended to combine tangrams with contextual information, without evoking unwanted inferences about, e.g., size and location relations, which might occur if the tangram were placed directly in or overlaying scene images.

\subsection{Data Collection}

We collect our data using the Argilla framework\footnote{https://argilla.io/} with crowdworkers from Prolific ($n=110$). 
Annotators are instructed to locate the tangram and describe what kind of object it depicts. With this, we ensure that participants pay attention to the scene images at least briefly while locating the tangram, given preceding work which indicates that the \textit{gist} of a scene is processed very quickly, cf. \citealt{Oliva2006}.
We collect 10 annotations per item, i.e., a total of $4070$ annotation points. Every annotator is assigned 37 items, which include exactly one item for each tangram in our data. 
100 annotators cover \textit{scene} conditions, i.e., tangram images are paired with random scene categories. Separate from this, 10 annotators provide descriptions for the \textit{baseline} condition, i.e., tangrams are coupled with uniform color patches. Workers are paid according to the local minimum wage.

\subsection{Post Processing}

We process the collected annotations using a combination of automatic tools and human validation or refinement. Every item in our dataset contains four annotations (\textit{raw}, \textit{label}, \textit{synset}, \textit{normalized label}), which are derived as follows:
First, we reduce the \textit{raw} annotations to \textit{labels}, i.e., nouns or compounds that denote the type of object the tangram is thought of depicting. If annotators provide more than one interpretation for a given tangram, we select only the first. 
After this, we map the labels to WordNet \textit{synsets} and select the first lemma from each synset as the \textit{normalized label} with reduced onomasiological variation.
We use spaCy \citep{Honnibal2020} and NLTK \citep{Bird2009} for label extraction and WordNet mapping. 
Both steps are manually validated and corrected.

\section{Data Analysis}

\subsection{Research Questions and Hypotheses}

In our analysis, we investigate whether people categorize and name tangrams differently if they co-occur with images of different scenes, i.e., if scene context affects the choice between alternative conceptual perspectives in tangram descriptions. 
Previous work has shown that tangrams vary in their \textit{conceptual flexibility}, i.e., there are different degrees of naming consensus for different tangram shapes, cf. \citealt{Ji2022}. For scene context effects on tangram descriptions, we expect the following patterns:

\begin{itemize}
    \item[H1] Scene context affects variation in tangram descriptions.
    \item[H2] Tangram descriptions elicited in context will be conceptually more coherent with this context as compared to descriptions elicited out of context.
    \item[H3] Context effects are more pronounced for certain combinations of tangrams and scenes, i.e. tangrams vary in their \textit{conceptual compatibility} to certain scenes.
\end{itemize}
    
\subsection{Analysis methods}

\begin{table*}
    \centering
    \small
    \begin{tabular}{l|ccc|cc}
  \toprule
  {}          & \multicolumn{3}{c|}{Variation} &   \multicolumn{2}{c}{\textsc{KiloGram} comparison}            \\
  scene       & SND mean/std                         &  SND corr. / \textsc{KiloGram}           & \% Top mean/std                         &   overlap            &  MRR                          \\
  \midrule           
  bathroom    & 0.92±0.09                         & 0.42***                                     &  27.0±16.1                              &    37.4              &               0.28            \\
  beach       & 0.92±0.13                         & 0.46***                                     &  26.8±17.2                              &    38.3              &               0.29            \\
  bedroom     & 0.93±0.11                         &  0.38**                                     &  25.4±14.8                              &    40.0              &               0.28            \\
  forest      & 0.94±0.06                         & 0.51***                                     &  25.4±11.2                              &    \textbf{40.3}     &               0.28            \\
  kitchen     & 0.92±0.11                         &  0.37**                                     &  \textbf{27.8}±17.0                     &    34.6              &               0.28            \\
  mountain    & 0.92±0.13                         & 0.41***                                     &  24.9±12.4                              &    39.2              &               0.29            \\
  office      & 0.91±0.14                & 0.51***                                     &  25.7±18.2                     &    35.0              &               0.26            \\
  sea\_bottom & \textbf{0.95}±0.05    &  0.47***                                    &  \underline{22.2}±8.5       &    35.5              &               \underline{0.24}\\
  sky         & 0.91±0.13                         &  \underline{0.33**}                         &  27.0±16.3                              &    \underline{34.0}  &               0.29            \\
  street      & 0.93±0.10                         & 0.51***                                     &  25.4±14.5                              &    37.9              &               0.27            \\
  \midrule
  none        & \underline{0.91}±0.13             & \textbf{0.61***}                            & 27.6±18.0                               &    38.7              &               \textbf{0.30}   \\
  \bottomrule
\end{tabular}
    \caption{Variation and overlap results for human annotations. SND correlations, overlap, and Mean Reciprocal Rank (MRR) are calculated with respect to annotations and scores for the same tangrams in \textsc{KiloGram}, asterisks denote significance levels with Kendall's tau (*p<0.05; **p<0.01; ***p<0.001). \textbf{Highest} and \underline{lowest} scores are highlighted.}
    \label{tab:variation_and_overlap}
\end{table*}

\paragraph{Shape Naming Divergence (SND) and \% Top}

We rely on SND \citep{Ji2022} and the inter-annotator agreement (\% Top, \citealt{silberer-etal-2020-object}) to estimate the degree of variation in our data. 
SND quantifies variability between annotations by measuring if tokens are used in multiple descriptions for the same item or are specific to individual descriptions. Following \citealt{Ji2022}, we use SND to analyze phrase-level tangram descriptions. To this end, we calculate SND scores for each set of raw descriptions for individual items.
We also correlate SND scores in our data with SND in \textsc{KiloGram}, using Kendall’s tau \citep{Kendall1938}.
\% Top is calculated for normalized object labels, i.e., lemmas of extracted synsets, by obtaining the relative frequency of the most frequent label for each item.

\paragraph{Lexical Overlap with \textsc{KiloGram} and Mean Reciprocal Rank (MRR)}

We use annotations from the \textit{dense} split in \textsc{KiloGram} as a benchmark for our data. Since we are particularly interested in the conceptualization of tangrams in terms of depicted object types, we extract the object labels from the phrases of the \textsc{KiloGram} annotations using spaCy. We then calculate the lexical overlap, i.e. the proportion of unique labels in our annotations that are also found in the \textsc{KiloGram} annotations for the same tangram, and the MRR of labels in our data and \textsc{KiloGram} ranked by frequency.

\paragraph{Label frequency}

To get interpretable estimates of scene effects on tangram descriptions, we compute the occurrence frequency of normalized object labels in the annotations, aggregating all tangram descriptions for each scene type. To identify context effects, we test the most frequent labels in all context conditions for significant deviation from the baseline condition using a chi-squared test.

\paragraph{Label-Scene similarity}

To quantify conceptual coherence and shifts in our data, we analyze the similarities between tangram descriptions and the scene context in which they were elicited, and compare this to the similarities between scene contexts and the baseline annotations, which are elicited without meaningful context (see Figure \ref{fig:item_example}). We test text-image similarity using CLIP \citep{Radford2021}, and text-text similarity using GloVe \citep{Pennington2014} and ConceptNet Numberbatch (\citealt{Speer2017}, henceforth Numberbatch). 
With CLIP, we encode the textual object labels extracted from the annotations and the images used as scene context and compute similarities between the labels and the mean representations of all three scene images used in each scene condition. For GloVe and Numberbatch, we replace the image vector with embeddings for the respective scene category label. 
Following \citealt{hessel-etal-2021-clipscore, takmaz-etal-2022-less}, we report coherence between tangram descriptions and scenes as scaled cosine similarities, i.e., 
\begin{equation}
sim(\vec d,\vec s) = 2.5 * max(cos(\vec d,\vec s), 0)
\nonumber
\end{equation}
and 
\begin{equation}
\textsc{Coherence}(D,s) = \frac{\sum_{d \in D} sim(\vec d,\vec s)}{|D|}
\nonumber
\end{equation}
, where $\vec d$ and $\vec s$ are vector representations for tangram descriptions (labels) and scene contexts (images or labels), and $d$ is a single label in the set of annotated labels $D$ for an item.
We especially look at \textit{in-context coherence} $\textsc{Coherence}(D_{t_i, s_i}, s_i)$, that is coherence between descriptions for tangrams in a certain scene context and the respective scenes, and \textit{baseline coherence} $\textsc{Coherence}(D_{t_i, s_b}, s_i)$, i.e. coherence between tangrams described in the baseline condition $s_b$ and certain scenes. To quantify conceptual shifts in scene context, we report the difference between these scores, that is
\begin{multline}
    \textsc{Shift}(i,s) = \\ 
    \textsc{Coherence}(D_{t_i, s_i}, s_i) - \textsc{Coherence}(D_{t_i, s_b}, s_i)
    \nonumber
\end{multline}
, representing the increase in scene coherence for the descriptions of an item $i~=~\langle t_i, s_i \rangle$ as compared to baseline descriptions of the same tangram. $\textsc{Shift} > 0$ indicates that tangrams are interpreted more coherently to the scenes they are placed in.

\subsection{Results}

\paragraph{SND and \% Top}

\begin{figure}
    \centering
    \includegraphics[width=1.0\columnwidth]{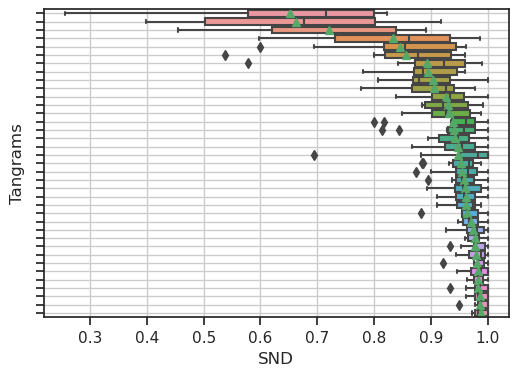}
    \caption{SND scores broken down for tangrams, markers indicate the mean. The overall distribution is skewed right, showing high variation in descriptions.
    }
    \label{fig:snd}
\end{figure}

\begin{table*}
    \footnotesize
    \centering
    \begin{tabular}{l|lllll}
\toprule
 & \#1 & \#2 & \#3 & \#4 & \#5 \\
 \midrule
bathroom & person (7.57 \%) & dog (3.78 \%) & \textbf{sink (3.24 \%)} & table (2.16 \%) & \textbf{wrench (2.16 \%)} \\
beach & person (5.41 \%) & dog (4.32 \%) & crab (3.51 \%) & horse (2.43 \%) & K (2.16 \%) \\
bedroom & person (6.49 \%) & dog (4.32 \%) & \textbf{bed (3.78 \%)} & \textbf{lamp (3.51 \%)} & table (2.16 \%) \\
forest & person (7.3 \%) & dog (4.05 \%) & bird (2.43 \%) & house (2.16 \%) & \textbf{forest (1.89 \%)} \\
kitchen & person (5.68 \%) & dog (5.68 \%) & \textbf{cabinet (2.7 \%)} & table (2.43 \%) & bird (2.43 \%) \\
mountain & \textbf{mountain (6.22 \%)} & person (5.14 \%) & bird (2.97 \%) & \textbf{dog (2.43 \%)} & \textbf{rock (1.89 \%)} \\
office & person (7.84 \%) & \textbf{desk (4.32 \%)} & table (3.24 \%) & dog (2.97 \%) & lamp (2.43 \%) \\
sea bottom & person (4.59 \%) & \textbf{fish (4.59 \%)} & turtle (2.43 \%) & \textbf{dog (2.43 \%)} & table (2.16 \%) \\
sky & person (5.41 \%) & bird (4.86 \%) & dog (4.05 \%) & \textbf{cloud (3.24 \%)} & mountain (2.43 \%) \\
street & person (7.57 \%) & dog (4.59 \%) & crab (2.7 \%) & house (1.89 \%) & bird (1.89 \%) \\
\midrule
none & person (7.03 \%) & dog (5.68 \%) & horse (2.97 \%) & bird (2.43 \%) & snake (2.16 \%) \\
\bottomrule
\end{tabular}
    \caption{Occurrence frequencies for WordNet Lemmas in the annotations for all tangrams in the given scene. Labels printed \textbf{bold} deviate significantly from the occurrence frequencies in the baseline condition (\textit{none}).}
    \label{tab:occ_freq}
\end{table*}

Aggregated over scenes, annotations in the baseline condition show the lowest SND, although differences between conditions are marginal (Table \ref{tab:variation_and_overlap}). Hence, on average, descriptions for tangrams show slightly more variation if paired with scene context, indicating that scenes can bring in additional ways of interpreting tangram shapes that are less accessible without context. 
Correlating the SND scores for different scene conditions with the scores provided in \textsc{KiloGram} reveals significant correlations in all cases, i.e., similar variance patterns for the same tangrams with or without context. However, the highest correlation can be seen for the baseline condition, whereas patterns in certain scene conditions deviate more from \textsc{KiloGram} (Table \ref{tab:variation_and_overlap}). This again points to general context effects at the level of variation patterns, supporting H1 in this regard.

Aggregating scores over tangrams shows large differences between shapes, in line with \citet{Ji2022}. However, mean SND scores are mostly close to the upper bound, indicating rich variation (Figure \ref{fig:snd}). Interestingly, lower scores tend to be associated with higher variance, suggesting that for the same tangrams, annotators are more aligned in certain scenes than in others.

The \% Top scores, calculated for labels normalized via WordNet, generally resemble the findings of the SND analysis (Table \ref{tab:variation_and_overlap}). Broken down for scenes, the high agreement can be seen in cases where SND scores indicate low variation (e.g., \textit{kitchen} or the baseline condition), and low agreement aligns with high variation (\textit{sea bottom}).

\paragraph{Lexical Overlap and MRR}

The overlap scores between labels in our annotations and the \textit{dense} split in \textsc{Kilogram} reveal no clear patterns, with the highest overlap in the \textit{forest} and the lowest in the \textit{sky} condition (see Table \ref{tab:variation_and_overlap}). However, our data still contains a high proportion of labels not included in the extensive \textsc{KiloGram} annotations, highlighting the high variability in tangram descriptions.
Taking into account the counts and frequency ranks of labels in both datasets, the MRR results in Table \ref{tab:variation_and_overlap} reveal similar patterns as SND and \% Top: Labels in the baseline condition resemble \textsc{KiloGram} annotations the most, and scores are generally lower with scene context.

\paragraph{Label frequency}
\label{sec:lex_freqs}

In Table \ref{tab:occ_freq}, we show the five most frequent object labels in all tangram descriptions for each scene condition, normalized via WordNet. Although certain labels are frequent in all conditions (e.g. \enquote{person}, \enquote{dog}), others occur more frequently in conceptually related scenes. In particular, most labels whose frequency deviates significantly from the baseline condition ($p < 0.05$ in the chi-squared test) denote objects typically occurring in the respective scenes, e.g. \enquote{sink} / \textit{bathroom}, \enquote{bed} / \textit{bedroom}, \enquote{cabinet} / \textit{kitchen} and, most prominently, \enquote{mountain} / \textit{mountain}. 
This supports our hypothesis H2, i.e., the significantly higher frequency of conceptually related labels indicates preferences towards tangram conceptualizations coherent with scene context.

\paragraph{Label-Scene similarity}

\begin{figure*}
    \centering
    \begin{subfigure}{0.31\textwidth}
        \includegraphics[width=\columnwidth]{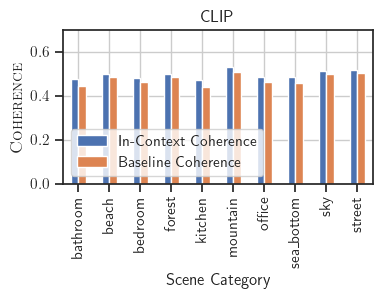}
    \end{subfigure}
    \begin{subfigure}{0.31\textwidth}
        \includegraphics[width=\columnwidth]{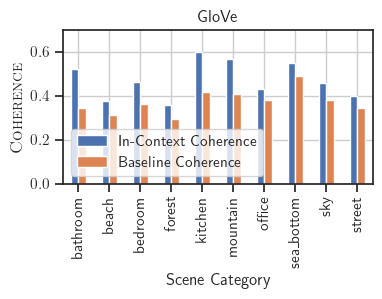}
    \end{subfigure}
    \begin{subfigure}{0.31\textwidth}
        \includegraphics[width=\columnwidth]{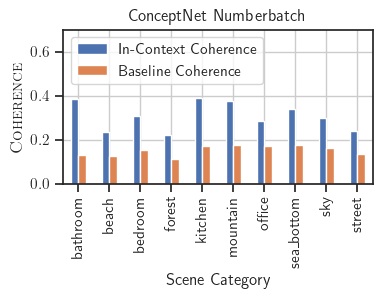}
    \end{subfigure}
    \caption{Mean \textsc{Coherence} scores between tangram annotations and scenes (images for CLIP, labels for GloVe and Numberbatch) for all scene categories. \textit{In-context coherence} consistently surpasses \textit{baseline coherence}, indicating that scene context causes semantically related conceptualizations in descriptions.}
    \label{fig:vector_similarities}
\end{figure*}

\begin{figure}
    \centering
    \includegraphics[width=1.0\columnwidth]{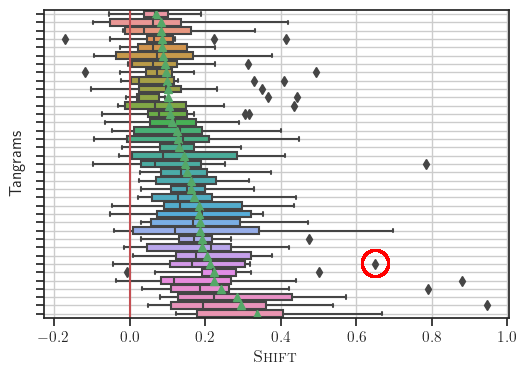}
    \caption{\textsc{Shift} scores broken down for tangrams. $\textsc{Shift} >0$ (red line) indicate that scene context leads to coherent tangram conceptualizations. The marked item is displayed in Figure \ref{fig:item_example} (\textit{bathroom} condition).}
    \label{fig:sim_deltas}
\end{figure}

The results of the conceptual coherence analysis are illustrated in Figure \ref{fig:vector_similarities}. Across all encoding methods and scenes, \textit{in-context} coherence surpasses \textit{baseline} coherence, i.e., labels that are elicited in a given scene context show higher similarity to corresponding scene representations than baseline annotations for the same tangrams. Generally, this indicates that annotators tend to produce descriptions for tangrams that align with scenes displayed to them, supporting H2.

Interestingly, the degree of conceptual shift depends on the embedding space used to encode labels and scenes. 
CLIP, which scores visual similarities, predicts smaller shifts than GloVe and Numberbatch, which encode more generic and conceptual similarities. This supports the interpretation that conceptual shifts triggered by scene context represent genuine conceptual variation and changes in perspective, rather than mere visual associations.
The largest differences can be seen with Numberbatch embeddings, possibly as a result of the explicit semantic relations included in the underlying ConceptNet meaning representation.

To illustrate conceptual shifts for individual tangrams, Figure \ref{fig:sim_deltas} shows the \textsc{Shift} scores for all tangram/scene combinations, computed with Numberbatch. For each tangram, a single data point represents the conceptual \textsc{Shift} with respect to a single scene, i.e., the difference between \textit{in-context} and \textit{baseline} coherence for this combination of tangram and scene.
Again, there is a clear trend towards annotations that are coherent to the respective context, i.e. $\textsc{Shift} > 0$ in the majority of cases. However, the degree of conceptual shifts varies between tangrams: whereas for some tangrams, descriptions in context show marginally higher scene coherence than baseline annotations, conceptualizations of other shapes seem to be more adaptable to different scenes. In particular, right outliers in this graph mark individual scenes for which tangram annotations elicited in context show a much higher similarity than the baseline annotations. This can be seen as cases where tangrams are conceptually compatible with certain scenes, with regard to interpretations that are less accessible without context, supporting hypothesis H3. 

\paragraph{Qualitative Examples}

Figure \ref{fig:item_example} shows labels for a single tangram in three scene conditions (\textit{bathroom}, \textit{beach}, and \textit{none}). With \textit{bathroom} context, this item has one of the highest \textsc{Shift} scores in our data (0.65, cf. Figure \ref{fig:sim_deltas}), as it includes a high rate of conceptually related labels (\enquote{sink}, \enquote{bathtub}), none of which occurs in the baseline condition. For \textit{beach}, annotations also have high relatedness to the scene, but lower \textsc{Shift} (0.17) since related labels are also included in the baseline condition.

\begin{table}[]
    \footnotesize
    \centering
    \begin{tabular}{l|c|cc}
      \toprule
                &         & \multicolumn{2}{c}{Overlap}               \\
      system    &  \% Top & \textsc{SceneGram} & \textsc{KiloGram}\\
      \midrule
      LLaVA-7b  &  58.50  & 26.61        & 34.99            \\
      LLaVA-13b &  36.71  & 21.13        & 34.42            \\
      LLaVA-34b &  59.17  & 27.64        & 50.22            \\
      LLaVA-72b &  79.46  & 26.00        & 54.52            \\
      \bottomrule
    \end{tabular}
    \caption{\% Top and \% Overlap with our human annotations (same item, normalized labels) and the \textsc{KiloGram} annotations (same tangram), global mean.}
    \label{tab:model_results}
\end{table}

\begin{figure*}
    \centering

    \begin{subfigure}{0.45\textwidth}
        \includegraphics[width=\columnwidth]{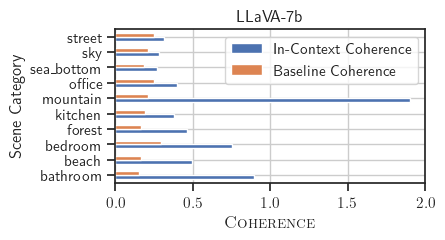}
    \end{subfigure}
    \begin{subfigure}{0.45\textwidth}
        \includegraphics[width=\columnwidth]{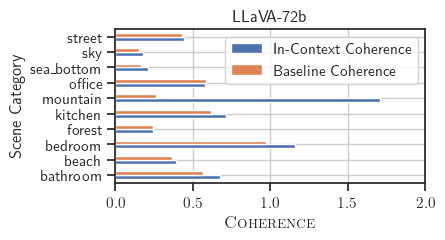}
    \end{subfigure}
    
    \caption{\textsc{Coherence} scores (Numberbatch) for LLaVA 7b (smallest) and 72b (largest). LLaVA 72b shows the higher overall scores (especially baseline coherence), but 7b has higher differences between in-context and baseline coherence. In both cases, in-context coherence spikes in the \textit{mountain} condition.
    }
    \label{fig:llava_cn_sim}
\end{figure*}

\section{Modelling Experiments}

Analyzing human annotations has shown that scene context affects tangram descriptions, i.e., tangrams are often interpreted in ways that align with the scenes they are placed in. 
In this section, we explore the tangram descriptions in context generated by off-the-shelf multimodal LLMs. 
For this, we generate tangram descriptions using the 7b, 13b, 34b and 72b parameter variants of LLaVA-NeXT \citep{Liu2023, Liu2024}\footnote{accessed via \href{https://huggingface.co/collections/llava-hf/llava-next-65f75c4afac77fd37dbbe6cf}{huggingface}, the latter using Int-8 quantization due to resource limitations.}, and test the outputs for variation, alignment with human data and conceptual shifts, using methods from the preceding analysis.

\subsection{Method}
\label{sec:modeling_method}
We use a two-step inference process to collect sets of tangram descriptions:
First, we prompt our models to predict the location of the tangram in the item grid, i.e., \textit{top/bottom} and \textit{left/right}.
After this, akin to \citet{testoni-etal-2024-naming} and keeping the location prediction as context, we repeatedly prompt our systems to generate descriptions of the tangram using nucleus decoding \citep{Holtzman2019} with \textit{p=0.5}. To facilitate subsequent analysis, we restrict responses to WordNet lemmas, i.e., after each response, we use SpaCy to extract the head noun from the generated description, assert that it is included in WordNet, and repeat the prediction process a maximum of 10 times if this is not the case. Inference terminates after 10 valid responses.

We calculate the relative frequency of the most frequent label in each response set (\% Top) to assess the variation in generated descriptions, and the overlap to extracted labels in \textsc{KiloGram} \textit{dense} and our data. Finally, we use \textsc{Coherence} scores to test for conceptual shifts. Accuracies for location predictions and frequency tables for scene conditions are included in the appendix.
    
\subsection{Results}

\paragraph{\% Top}

As we repeatedly sample the output distributions of the same models,\% top indicates if the systems are able to conceptualize the same tangrams in different ways. 
The results in Table \ref{tab:model_results} show that apart from LLaVA 13b, scores increase with size, i.e., models converge on a limited set of interpretations per item. Although the exact scores depend on decoding parameters, we note that on average, the most frequent labels occur much more often than in the human data (between 20 and 30~\%, see Table \ref{tab:variation_and_overlap}), indicating that individual systems cannot capture the range of human tangram conceptualizations.

\paragraph{Lexical Overlap}

With less than 30 \% of the generated labels also occurring in human annotations for the same items, the overlap between system responses and annotations in our data is surprisingly small (Table \ref{tab:model_results}). This indicates that system predictions seldom coincide with labels produced by humans, raising doubts about their general capability of replicating human-like conceptualization of abstract depictions.
However, overlap with \textsc{KiloGram} is higher, possibly due to their extensive annotations ($n \geq 50$ per item).

\paragraph{Label-Scene Similarity}

The \textsc{Coherence} scores in Figure \ref{fig:llava_cn_sim} suggest that our systems generate labels with higher coherence to 
scenes if tangrams are paired with the respective images, similar to human annotations.
However, coherence scores and differences between baseline and in-context predictions are often much higher, especially for \textit{mountain} scenes. 
Occurrence counts of generated labels show that here, systems generate the label \enquote{mountain} in up to 74 \% of cases (cf. Appendix \ref{app:generated_frequencies}), raising doubts about whether models rather describe scenes than tangrams, i.e., fail to parse the visually complex items.

\section{Discussion and Conclusion}

The role of context in the conceptualization of visual objects and its interactions with variability and creativity in referring are notoriously hard to grasp for (computational) linguistic research.
\textsc{SceneGram} addresses this gap, proposing a controlled paradigm that elicits descriptions of conceptually ambiguous tangram shapes in scene context.
Our results underpin a common theoretical assumption that has, however, been rarely tested ``in the wild'', especially not in language \& vision research: scene context can fundamentally shape speakers' conceptual perspectives when describing a visual stimulus.
\textsc{SceneGram} shows that tangram descriptions in context remain highly diverse while becoming conceptually more coherent with the scene context. 
Experiments with off-the-shelf multimodal LLMs indicate that the systems cannot reproduce human variance and conceptualizations of tangrams, but demonstrate general effects of scene context. 
Overall, our results highlight the importance of scene context on object naming at the level of conceptualization, pointing to weaknesses of current multimodal language models in this regard. 

For future work, our data can be used to, e.g., probe more general mechanisms of visuo-linguistic processing in multimodal LLMs, similar to work in cognitive science or psycholinguistics, where using abstract visual stimuli is a well-established paradigm.
At the same time, our approach also raises new questions and directions. 
More work is needed to understand how exactly context modulates the variance of conceptualizations. Data in \textsc{SceneGram} suggests that context could \textit{prime} humans for certain interpretations, effectively reducing variance, or could evoke entirely new interpretations without blocking preexisting ones, pointing to interesting connections to creativity.
Further work is also needed to understand how conceptual perspective and scene context interact when the communicative effectiveness of object descriptions is at stake, e.g. in a reference game. Here, the creative use of language is a necessity in the first place. 
Combining abstract stimuli with contextual information or communicative demands could be a valuable tool for future research to study creativity and linguistic variability in humans and language models.

\section*{Limitations}

We identify the following limitations in our study:

First, we note that the annotation procedure could be further refined. In particular, more advanced setups with defined communicative objectives as in e.g. reference games could further ensure high-quality descriptions for tangrams and elicit tangram descriptions that are not only creative, but also effective if communication. Our primary interest here is to collect and analyze general descriptions for tangrams in varying scene contexts, but we see potential for further insights by adapting and improving our methods.

Second, the robustness of our findings could be further improved by scaling up the data collection. By focusing on a subset of the tangrams in \textsc{KiloGram}, we were able to achieve an acceptable sample size of 10 annotations for each combination of tangrams and scenes within our financial and time limits. 
However, due to the high variance in our data, a larger pool of annotations per item could allow for more reliable or comprehensive conclusions.

Finally, for the modelling experiments, further system architectures and hyperparameter configurations could be added for more comprehensive insights, possibly including commercial systems such as ChatGPT, Gemini, DeepSeek and Claude. Due to space and time constraints, we leave this for future research. 

\section*{Acknowledgments}

This research has been funded by the \href{https://www.dfg.de/}{Deutsche Forschungsgemeinschaft} (DFG, German Research Foundation) -- \href{https://gepris.dfg.de/gepris/projekt/512393437}{CRC-1646, project number 512393437}, project B02.

\bibliography{bib_export}

\appendix

\section{Crowdsourcing Procedure}

\begin{figure*}
    \includegraphics[width=\textwidth]{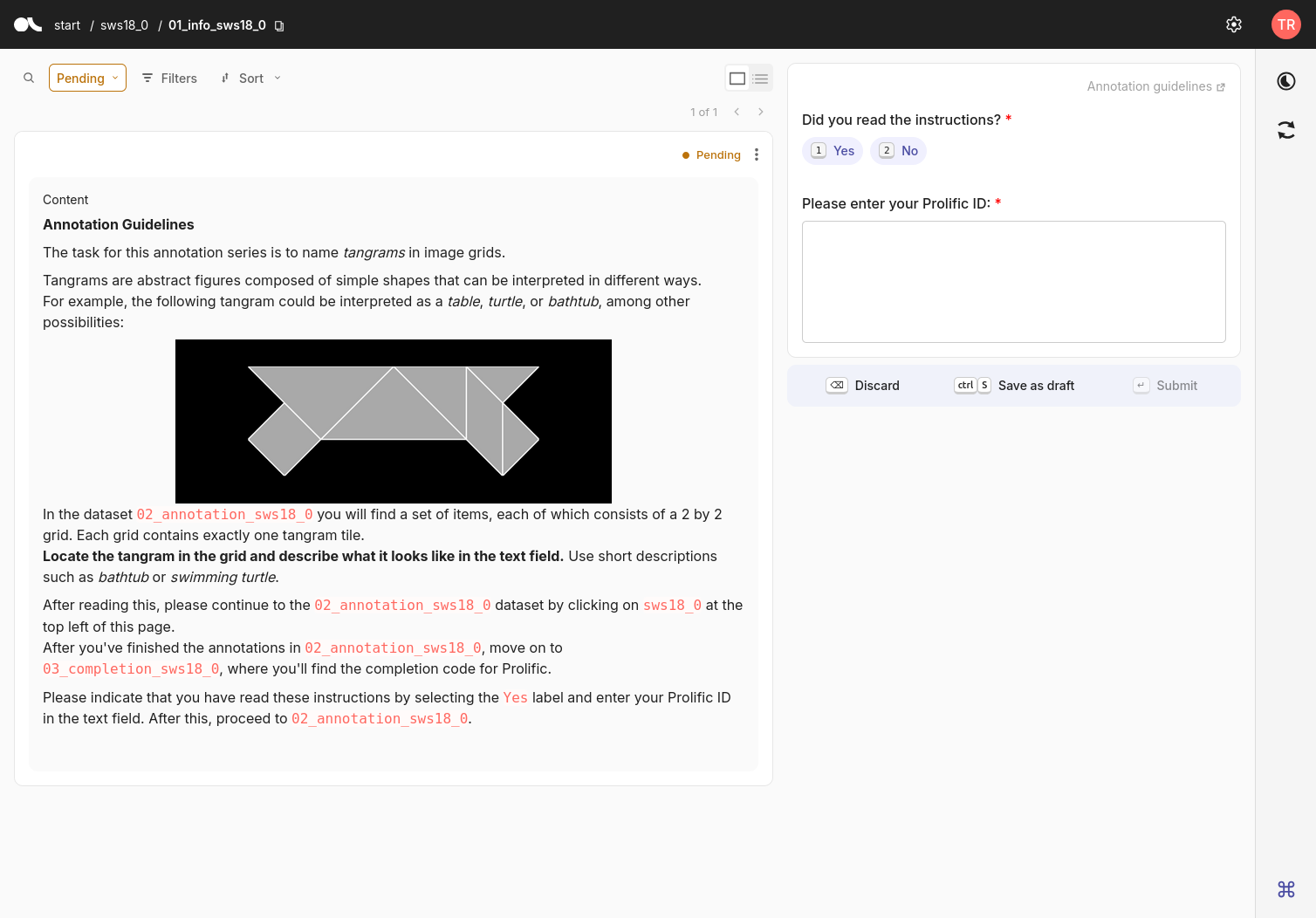}
    \caption{Screenshot of the annotation instructions}
    \label{fig:ann_prompt}
\end{figure*}

\begin{figure*}
    \includegraphics[width=\textwidth]{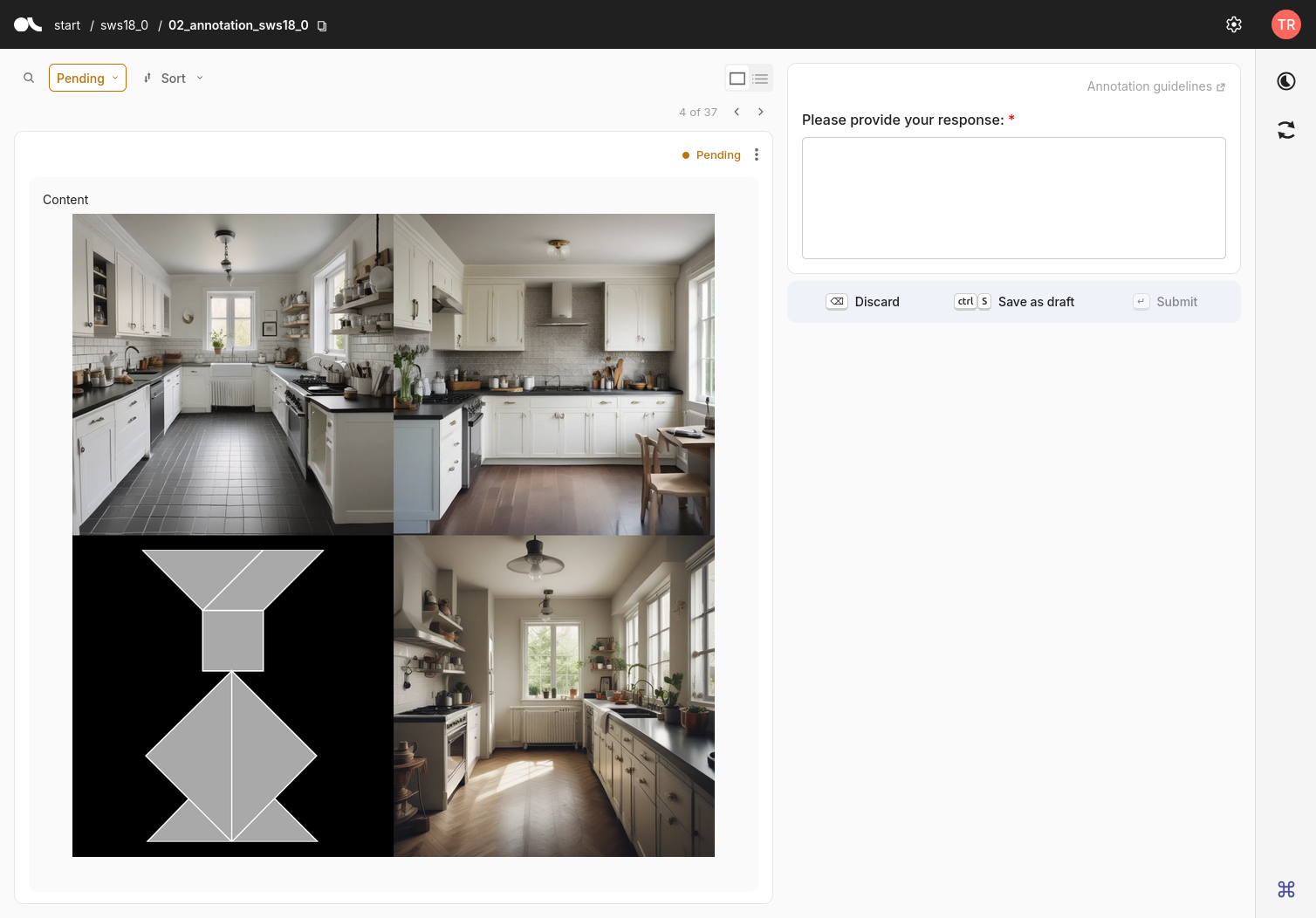}
    \caption{Screenshot of the annotation setup}
    \label{fig:ann_setup}
\end{figure*}

For our data collection, we recruited 110 annotators, all of which are located in the United States and have stated English as their primary language. Workers were paid according to the local minimum wage, and the intended purpose of the data was explained. We recruited the crowdworkers via Prolific, and used Argilla (hosted on Huggingface Spaces) to collect annotations. 
The annotators are heterogeneous in terms of their 
age (18-29: 25.3\%; 30-39: 34.3\%; 40-49: 22.2\%; 60+: 18.2\%), 
ethnicity (White: 58.6\%; Black: 18.2\%; Asian: 11.1\%; other: 12.1\%)
and sex (female: 51.5\%; male: 48.5\%).
A screenshot of the annotation instructions can be seen in Figure \ref{fig:ann_prompt}. The annotation interface can be seen in Figure \ref{fig:ann_setup}.

To validate the automatic processing steps, we provided student assistants with the automatically processed data and instructed them to manually check whether the correct labels were extracted and valid WordNet synsets were selected, and update labels and synsets if necessary.

\section{Scientific Artifacts}
\label{app:scientific-artifacts}

In our work, we mainly use scientific artifacts in the form of publicly available datasets and model implementations, as well as Python frameworks and modules (cf. Appendix~\ref{sec:implementation_details}).
In all cases, we are confident that our work is consistent with their intended use. Most importantly, our work builds on the \textsc{KiloGram} dataset as a source of tangram figures. The dataset is available on \href{https://github.com/lil-lab/kilogram}{GitHub}. 
To generate scene images we rely on \textit{SDXL-Lightning} \citep{Lin2024}, accessed via the provided \href{https://huggingface.co/spaces/ByteDance/SDXL-Lightning}{Huggingface Space}. The model is available on \href{https://huggingface.co/ByteDance/SDXL-Lightning}{huggingface}.
Our data and code for this project are available at \href{https://github.com/clause-bielefeld/scenegram}{github.com/clause-bielefeld/scenegram}.

\section{Risks and Ethical Considerations}

We do not believe that there are significant risks associated with this work, as we work with descriptions of abstract items which are not believed to be perceived as hurtful, and release data with limited scale. No ethics review was required. Our data does not contain any protected information and is fully anonymized.

\section{Prompts and Model Inference}

Our model prompts consist of two parts. First, we instruct the systems to predict the locations of tangrams in the item grids, using greedy decoding:

\enquote{In this 2 by 2 grid, exactly one tile contains a tangram figure. In which grid cell is it? Pick your response from the following options: Top left, top right, bottom left, bottom right.}

After this, keeping the location prompt and predictions as context, we instruct the models to generate tangram descriptions:

\enquote{Describe what this tangram looks like. Ignore the other tiles. Keep your answer short and concise. Give your answer in the form: The tangram depicts a \_.}

As described in Section \ref{sec:modeling_method}, we repeatedly prompt our systems using the last prompt, to collect a diverse set of responses (using nucleus decoding with $p=0.5$). For each generated response, we extract the head noun from the model response using spaCy, assert that it is included in WordNet, and repeat the prediction for a maximum of 10 times if this is not the case. Inference terminates after 10 valid responses.

\section{Implementation Details}
\label{sec:implementation_details}

For our experiments we rely on models from \href{https://huggingface.co/}{huggingface}. In detail, we used the following models:

\begin{itemize}
    \item \href{https://huggingface.co/llava-hf/llava-v1.6-vicuna-7b-hf}{llava-hf/llava-v1.6-vicuna-7b-hf}
    \item \href{https://huggingface.co/llava-hf/llava-v1.6-vicuna-13b-hf}{llava-hf/llava-v1.6-vicuna-13b-hf} 
    \item \href{https://huggingface.co/llava-hf/llava-v1.6-34b-hf}{llava-hf/llava-v1.6-34b-hf}
    \item \href{https://huggingface.co/llava-hf/llava-next-72b-hf}{llava-hf/llava-next-72b-hf} (quantized using the \emph{bitsandbytes} library)
\end{itemize}

To generate responses with our models, we used Python 3.9.20 with the following libraries:
torch (2.5.1), transformers (4.46.2), bitsandbytes (0.44.1).
For data analysis we used Python 3.10.9, mostly using the following frameworks:
nltk (3.8.1), numpy (1.23.5), pandas (1.5.2), scikit-learn (1.2.0), scipy (1.9.3), seaborn (0.12.2), spacy (3.5.3 and the en\_core\_web\_sm model).

We used three NVIDIA RTX A6000 GPUs for inference for LLaVA 72b, two GPUs of the same type for LLaVA 34b and a single GPU of the same type for the remaining models. Depending on model size, generating responses took between 50 min (LLaVA 7b) and 7h (LLaVA 72b).

\section{Further Examples}

\begin{figure*}
    \footnotesize
    \centering
    \begin{tabular}{p{0.3\textwidth}p{0.3\textwidth}p{0.3\textwidth}}
        \centering\arraybackslash \includegraphics[width=.22\textwidth]{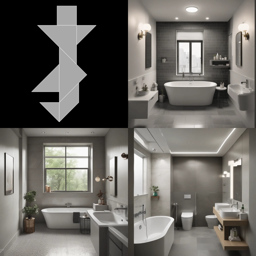} &
        \centering\arraybackslash \includegraphics[width=.22\textwidth]{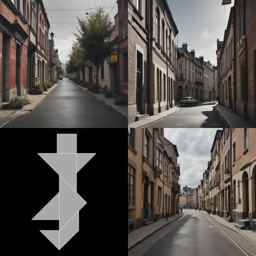} &
        \centering\arraybackslash \includegraphics[width=.22\textwidth]{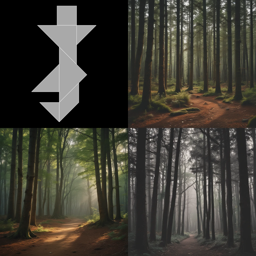} \\
        \textbf{human}: wrench (5); pipe wrench (2); hook (1); building (1); faucet (1)
        & \textbf{human}: streetlamp (1); bird (1); drill (1); screw (1); spike (1); street sign (1); rose (1); bird talon (1); wrench (1); person (1)
        & \textbf{human}: monkey wrench (1); deer (1); tree (1); fishhook (1); wrench (1); sword (1); person (1); cactus (1); totem pole (1); chicken (1)\\
        \textbf{LLaVA 7b}: bathtub (3); chair (2); house (2); triangle (2); diamond (1) &
        \textbf{LLaVA 7b}: diamond (3); house (3); shape (1); triangle (1); figure (1); staircase (1) &
        \textbf{LLaVA 7b}: diamond (5); house (2); triangle (2); figure (1) \\
        \textbf{LLaVA 72b}: house (6); cross (2); cat (1); bird (1)&
        \textbf{LLaVA 72b}: bird (5); cross (3); house (2)&
        \textbf{LLaVA 72b}: house (6); bird (3); person (1)\\
    \end{tabular}

    \begin{tabular}{p{0.3\textwidth}p{0.3\textwidth}p{0.3\textwidth}}
        \centering\arraybackslash \includegraphics[width=.22\textwidth]{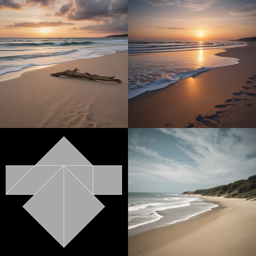} &
        \centering\arraybackslash \includegraphics[width=.22\textwidth]{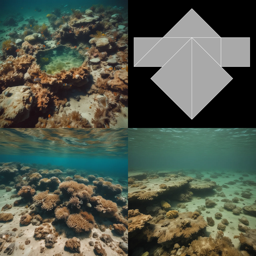} &
        \centering\arraybackslash \includegraphics[width=.22\textwidth]{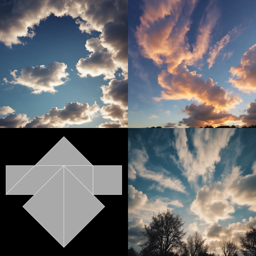} \\
        \textbf{human}: turtle (2); sailboat (1); wing suit (1); origami (1); airplane (1); kite (1); box kite (1); tie (1); map (1)
        & \textbf{human}: stingray (2); person (1); radio (1); iceberg (1); turtle (1); pyramids (1); water fountain (1); sea ray (1); duck (1)
        & \textbf{human}: pyramids (1); monkey face (1); aeroplane (1); kite (1); shell (1); airplane (1); stealth bomber (1); tissue box (1); stingray (1); spaceship (1)\\
        \textbf{LLaVA 7b}: beach (5); sunset (2); triangle (2); sun (1) &
        \textbf{LLaVA 7b}: diamond (6); square (2); hexagon (2) &
        \textbf{LLaVA 7b}: diamond (5); square (2); cloud (2); triangle (1) \\
        \textbf{LLaVA 72b}: house (10)&
        \textbf{LLaVA 72b}: house (10)&
        \textbf{LLaVA 72b}: house (10)\\
    \end{tabular}

    \begin{tabular}{p{0.3\textwidth}p{0.3\textwidth}p{0.3\textwidth}}
        \centering\arraybackslash \includegraphics[width=.22\textwidth]{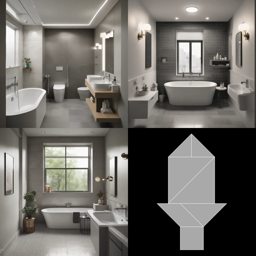} &
        \centering\arraybackslash \includegraphics[width=.22\textwidth]{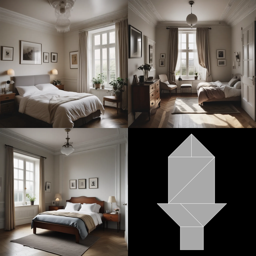} &
        \centering\arraybackslash \includegraphics[width=.22\textwidth]{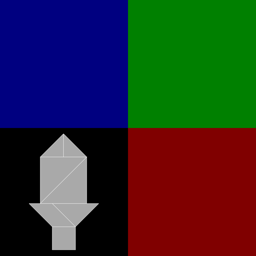} \\
        \textbf{human}: toilet (2); lightbulb (1); fountain (1); sink (1); bathtub (1); brush (1); water fountain (1); rocketship (1); rose (1) 
        & \textbf{human}: pot (1); lighthouse (1); light (1); candle (1); torch (1); dagger (1); vase (1); house (1); lamp (1); mirror (1)
        & \textbf{human}: candle (2); library (1); lighthouse top (1); sword (1); torch (1); pot (1); corn (1); flower bud (1); altar (1)\\
        \textbf{LLaVA 7b}: house (5); triangle (3); bathtub (2) &
        \textbf{LLaVA 7b}: house (6); pyramid (3); diamond (1) &
        \textbf{LLaVA 7b}: pyramid (7); triangle (3) \\
        \textbf{LLaVA 72b}: house (7); pyramid (3)&
        \textbf{LLaVA 72b}: house (7); pyramid (2); triangle (1)&
        \textbf{LLaVA 72b}: house (6); pyramid (4)\\
    \end{tabular}
    
    \caption{Further examples with human annotations and model predictions. Humans often produce labels which are coherent to scenes; whereas models are considerably less creative.}
    \label{fig:more_examples}
\end{figure*}

Figure \ref{fig:more_examples} contains further qualitative examples for items with high \textsc{Shift} scores for human annotations.

\section{Location determination accuracies}

In Table \ref{tab:location_determination} we report accuracy scores for target locations in the item grids as predicted by our systems. The results indicate that most models handle the location task without major problems, with the exception of LLaVa 7b. Whereas all larger variants produce no or only singular mistakes, this system struggles especially in the baseline condition, where the remaining grid cells are filled with uniform colors. While this does not mean that it fails to describe the tangram, the low location determination scores point to parsing difficulties which should be taken into consideration.

\begin{table}[]
    \centering
    \footnotesize
    \begin{tabular}{l|rrrr}
\toprule
      & \multicolumn{4}{c}{LLaVA} \\
scene &  7b &  13b &  34b &  72b \\
\midrule
bathroom   &      86.5 &      100.0 &      100.0 &      100.0 \\
beach      &      83.8 &      100.0 &      100.0 &      100.0 \\
bedroom    &      78.4 &      100.0 &      100.0 &      100.0 \\
forest     &      89.2 &      100.0 &      100.0 &      100.0 \\
kitchen    &      75.7 &       97.3 &      100.0 &       97.3 \\
mountain   &      91.9 &      100.0 &      100.0 &      100.0 \\
office     &      83.8 &      100.0 &      100.0 &       97.3 \\
sea\_bottom &      73.0 &      100.0 &      100.0 &       94.6 \\
sky        &      83.8 &      100.0 &      100.0 &      100.0 \\
street     &      86.5 &      100.0 &      100.0 &      100.0 \\
\midrule
none       &      56.8 &      100.0 &      100.0 &      100.0 \\
\bottomrule
\end{tabular}
    \caption{Location determination accuracies (\%)}
    \label{tab:location_determination}
\end{table}

\section{Lexical Classes in Scene Contexts}

Using WordNet, we are able to abstract the label frequency analysis to more general sets of lexical items. For this, we map our annotation synsets to a pre-defined set of reference synsets (\textit{artifact.n.01}, \textit{animal.n.01}, \textit{person.n.01}, \textit{geological\_formation.n.01}, \textit{written\_symbol.n.01} and \textit{entity.n.01} as a generic fallback), selecting the reference synset with the highest distance from the WordNet root to which the annotated synset is a recursive hyponym as the lexical category. 

\begin{figure*}
    \centering
    \includegraphics[width=.8\linewidth]{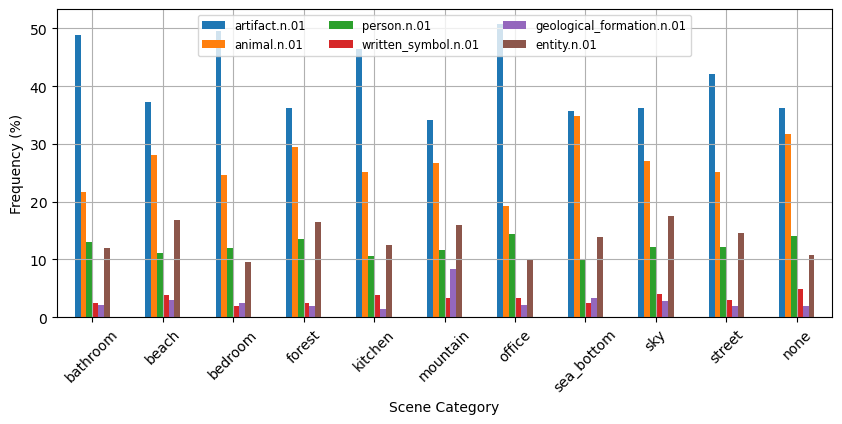}
    \caption{Occurrence frequencies for labels in different lexical classes for human annotations, calculated via WordNet.}
    \label{fig:lex_classes}
\end{figure*}

Figure \ref{fig:lex_classes} shows that rankings are similar between scene conditions, i.e., \textit{artifact} and \textit{animal} are ranked first and second in all cases. However, there are differences in the frequency of occurrence: While \textit{artifacts} is especially common for indoor scenes (\textit{bathroom}, \textit{bedroom}, \textit{kitchen}, \textit{office}), \textit{animals} is slightly more frequent in natural or outdoor scenes (\textit{beach}, \textit{forest}, \textit{mountain}, \textit{sea bottom}, \textit{sky}). \textit{geological\_formation}, with labels like \enquote{mountain} or \enquote{hill}, is especially frequent in the \textit{mountain} condition, reflecting our label frequency results.

\section{Label Frequencies for Generated Descriptions}
\label{app:generated_frequencies}

Table \ref{tab:model_label_frequencies} shows occurrence frequencies for labels in tangram descriptions as generated by LLaVA variants, aggregated over scenes.  

\begin{table*}
    \footnotesize
    \begin{subtable}{\textwidth}
        \begin{tabular}{l|lllll}
            \toprule
            {} &                  \#1 &                  \#2 &                  \#3 &                 \#4 &                  \#5 \\
            \midrule
            bathroom   &   bathtub (28.65 \%) &     house (19.19 \%) &    diamond (7.03 \%) &    person (6.76 \%) &  rectangl (5.95 \%) \\
            beach      &       sun (14.59 \%) &     house (12.43 \%) &      bird (10.27 \%) &  triangle (9.46 \%) &      beach (9.19 \%) \\
            bedroom    &     house (28.11 \%) &       bed (14.59 \%) &   diamond (10.54 \%) &      bird (8.38 \%) &       chair (7.3 \%) \\
            forest     &   diamond (14.32 \%) &     house (11.35 \%) &  triangle (10.27 \%) &    forest (8.65 \%) &       tree (7.57 \%) \\
            kitchen    &     house (28.65 \%) &      bird (12.16 \%) &    diamond (7.57 \%) &      chair (7.3 \%) &    triangle (7.3 \%) \\
            mountain   &  mountain (74.05 \%) &   triangle (4.05 \%) &     figure (3.78 \%) &    person (3.24 \%) &    pyramid (2.97 \%) \\
            office     &     chair (14.86 \%) &     house (14.05 \%) &    diamond (8.92 \%) &     person (7.3 \%) &    triangle (7.3 \%) \\
            sea bottom &   triangle (9.19 \%) &    diamond (8.65 \%) &     square (7.57 \%) &      fish (7.03 \%) &     letter (7.03 \%) \\
            sky        &  triangle (14.32 \%) &      bird (13.78 \%) &     house (11.35 \%) &  diamond (10.27 \%) &     letter (7.84 \%) \\
            street     &      house (22.7 \%) &  triangle (11.35 \%) &   diamond (10.81 \%) &    person (8.65 \%) &        dog (7.57 \%) \\
            \midrule
            none       &    diamond (17.3 \%) &     house (14.05 \%) &  triangle (10.54 \%) &    square (8.38 \%) &       bird (7.84 \%) \\
            \bottomrule
        \end{tabular}
        \caption{LLaVA 7b}
    \end{subtable}

    \begin{subtable}{\textwidth}
        \begin{tabular}{l|lllll}
            \toprule
            {} &                  \#1 &                \#2 &                \#3 &                \#4 &                  \#5 \\
            \midrule
            bathroom   &     house (18.11 \%) &    bird (16.22 \%) &  person (13.78 \%) &  square (10.81 \%) &       tree (9.19 \%) \\
            beach      &      bird (19.73 \%) &  person (14.05 \%) &   house (11.35 \%) &    tree (10.27 \%) &          a (7.57 \%) \\
            bedroom    &     house (20.81 \%) &    bird (17.84 \%) &   person (12.7 \%) &   square (7.84 \%) &          a (7.57 \%) \\
            forest     &      bird (26.22 \%) &  person (18.92 \%) &   house (15.14 \%) &     tree (8.65 \%) &     animal (4.86 \%) \\
            kitchen    &     house (20.81 \%) &    bird (18.65 \%) &  person (15.95 \%) &   square (8.38 \%) &       tree (5.41 \%) \\
            mountain   &  mountain (35.41 \%) &   house (16.76 \%) &     bird (12.7 \%) &  person (10.81 \%) &  landscape (6.49 \%) \\
            office     &     house (18.92 \%) &    bird (16.49 \%) &  person (14.86 \%) &  square (11.35 \%) &       tree (5.41 \%) \\
            sea\_bottom &      bird (21.35 \%) &  person (14.86 \%) &     fish (9.19 \%) &    house (8.92 \%) &     square (8.38 \%) \\
            sky        &      bird (24.86 \%) &   house (18.11 \%) &  person (14.32 \%) &   square (7.03 \%) &       tree (6.49 \%) \\
            street     &      bird (20.54 \%) &  person (18.11 \%) &   house (15.68 \%) &  square (11.35 \%) &          a (5.95 \%) \\
            \midrule
            none       &      bird (24.86 \%) &  person (19.19 \%) &   house (10.81 \%) &  square (10.27 \%) &      shape (6.76 \%) \\
            \bottomrule
        \end{tabular}
        \caption{LLaVA 13b}
    \end{subtable}

    \begin{subtable}{\textwidth}
        \begin{tabular}{l|lllll}
            \toprule
            {} &                  \#1 &                  \#2 &                  \#3 &                  \#4 &                \#5 \\
            \midrule
            bathroom   &      bird (25.14 \%) &     house (14.86 \%) &  triangle (14.32 \%) &        dog (8.92 \%) &      man (4.86 \%) \\
            beach      &      bird (24.05 \%) &  triangle (15.68 \%) &       dog (13.51 \%) &      house (10.0 \%) &      man (3.78 \%) \\
            bedroom    &      bird (20.54 \%) &     house (18.11 \%) &       dog (14.05 \%) &  triangle (12.97 \%) &    horse (4.05 \%) \\
            forest     &      bird (21.62 \%) &  triangle (19.73 \%) &       dog (10.81 \%) &      house (9.73 \%) &      man (4.32 \%) \\
            kitchen    &      bird (22.43 \%) &  triangle (16.49 \%) &     house (13.78 \%) &        dog (12.7 \%) &   square (4.05 \%) \\
            mountain   &  triangle (21.08 \%) &  mountain (16.22 \%) &      bird (12.16 \%) &        dog (8.92 \%) &    house (7.84 \%) \\
            office     &      bird (21.62 \%) &  triangle (17.57 \%) &       dog (11.62 \%) &     house (11.62 \%) &   person (5.68 \%) \\
            sea\_bottom &  triangle (20.27 \%) &      bird (19.46 \%) &         dog (7.3 \%) &      house (6.76 \%) &      man (4.59 \%) \\
            sky        &      bird (23.78 \%) &  triangle (17.03 \%) &        dog (10.0 \%) &      house (9.73 \%) &      man (5.95 \%) \\
            street     &  triangle (20.27 \%) &      bird (19.19 \%) &       dog (12.16 \%) &     house (11.62 \%) &  diamond (5.14 \%) \\
            \midrule
            none       &      bird (17.03 \%) &  triangle (14.86 \%) &       dog (11.08 \%) &      house (8.92 \%) &   person (3.78 \%) \\
            \bottomrule
        \end{tabular}
        \caption{LLaVA 34b}
    \end{subtable}

    \begin{subtable}{\textwidth}
        \begin{tabular}{l|lllll}
            \toprule
            {} &                  \#1 &               \#2 &               \#3 &                \#4 &                \#5 \\
            \midrule
            bathroom   &     house (71.08 \%) &    bird (6.76 \%) &  letter (5.14 \%) &   person (3.51 \%) &   figure (1.89 \%) \\
            beach      &     house (64.86 \%) &     bird (7.3 \%) &  letter (5.68 \%) &   person (5.14 \%) &      dog (3.51 \%) \\
            bedroom    &     house (69.73 \%) &  letter (6.22 \%) &  person (4.59 \%) &     bird (4.32 \%) &      bed (3.24 \%) \\
            forest     &     house (63.78 \%) &  person (8.38 \%) &     dog (5.14 \%) &     bird (4.59 \%) &   letter (3.51 \%) \\
            kitchen    &     house (68.92 \%) &  person (4.59 \%) &    bird (4.05 \%) &   letter (3.78 \%) &      dog (3.78 \%) \\
            mountain   &  mountain (64.86 \%) &  house (19.19 \%) &  person (5.95 \%) &  pyramid (2.43 \%) &   letter (2.16 \%) \\
            office     &      house (57.3 \%) &  person (7.84 \%) &  letter (7.84 \%) &     bird (4.86 \%) &  pyramid (3.24 \%) \\
            sea\_bottom &     house (54.32 \%) &     bird (7.3 \%) &    boat (6.76 \%) &   person (4.86 \%) &   letter (4.05 \%) \\
            sky        &     house (62.43 \%) &    bird (9.19 \%) &  person (5.14 \%) &    horse (4.59 \%) &   letter (3.51 \%) \\
            street     &     house (64.59 \%) &  person (8.11 \%) &   letter (7.3 \%) &     bird (5.41 \%) &       dog (2.7 \%) \\
            \midrule
            none       &     house (58.92 \%) &  person (6.49 \%) &    None (5.41 \%) &     bird (4.86 \%) &    horse (4.05 \%) \\
            \bottomrule
        \end{tabular}
        \caption{LLaVA 72b}
    \end{subtable}
    \caption{Occurrence frequencies for labels in tangram descriptions generated by LLaVA variants, aggregated over scenes.}
    \label{tab:model_label_frequencies}

\end{table*}

\end{document}